\documentclass[runningheads]{llncs}

\usepackage{eccv}

\usepackage{eccvabbrv}

\definecolor{cvprblue}{rgb}{0.21,0.49,0.74}
\usepackage{mathtools}
\usepackage{amsmath}
\usepackage{multirow}
\usepackage{wrapfig}
\usepackage{amssymb}
\usepackage{pifont}

\usepackage{color,colortbl}
\definecolor{ggray}{rgb}{0.92, 0.92, 0.99}

\usepackage{graphicx}
\usepackage{booktabs}
\usepackage{tabularx}

\usepackage[accsupp]{axessibility}  

\usepackage[pagebackref,breaklinks,colorlinks,citecolor=eccvblue]{hyperref}

\usepackage{orcidlink}

\DeclarePairedDelimiterX{\infdivx}[2]{(}{)}{%
  #1\;\delimsize\|\;#2%
}

\newcommand\norm[1]{\lVert#1\rVert}
\newcommand*\samethanks[1][\value{footnote}]{\footnotemark[#1]}

\begin{document}

\title{DreamDA: Generative Data Augmentation with Diffusion Models} 


\author{Yunxiang Fu\inst{1}\thanks{First two authors contributed equally.} \and
Chaoqi Chen\inst{1}\samethanks \and
Yu Qiao\inst{2} \and
Yizhou Yu\inst{1}
}

\authorrunning{Fu et al.}

\institute{Department of Computer Science, The University of Hong Kong \and
Shanghai Artificial Intelligence Laboratory \\
\email{yunxiangfu2001@gmail.com}, \email{cqchen1994@gmail.com}, \\ \email{qiaoyu@pjlab.org.cn}, \email{yizhouy@acm.org}}

\maketitle

\begin{abstract}
The acquisition of large-scale, high-quality data is a resource-intensive and time-consuming endeavor. 
Compared to conventional Data Augmentation (DA) techniques (e.g. cropping and rotation), 
exploiting prevailing diffusion models for data generation has received scant attention in classification tasks.
Existing generative DA methods either inadequately bridge the domain gap between real-world and synthesized images, or inherently suffer from a lack of diversity.  
To solve these issues, this paper proposes a new classification-oriented framework \textbf{DreamDA}, which enables data synthesis and label generation by way of diffusion models. 
DreamDA generates diverse samples that adhere to the original data distribution by considering training images in the original data as seeds and perturbing their reverse diffusion process.
In addition, since the labels of the generated data may not align with the labels of their corresponding seed images, we introduce a self-training paradigm for generating pseudo labels and training classifiers using the synthesized data.
Extensive experiments across four tasks and five datasets demonstrate consistent improvements over strong baselines, revealing the efficacy of DreamDA in synthesizing high-quality and diverse images with accurate labels. Our code will be available at
\url{https://github.com/yunxiangfu2001/DreamDA}.

\end{abstract}

\section{Introduction}

Over the past decade, deep learning has experienced a remarkable surge in success across a wide range of 
computer vision tasks. 
One of the main driving forces is the availability of large-scale and high-quality training datasets, such as ImageNet~\cite{russakovsky2015imagenet}, MS COCO~\cite{lin2014microsoft}, and LAION-5B~\cite{schuhmann2022laion}. 
However, high-quality large-scale data collection and annotation can be costly and time-consuming~\cite{alzubaidi2023survey}. 
This limits the effective deployment of deep learning in many applications as deep models demonstrate sub-optimal performance when training data is limited in size and diversity~\cite{qi2020small}.

To mitigate data scarcity, data augmentation (DA) techniques have been extensively explored.
Early DA methods apply simple transformations, random cropping, flipping, and color jittering, while more recent approaches (\emph{e.g.} mixup~\cite{zhang2017mixup} and CutMix~\cite{yun2019cutmix}) 
provide additional data by transforming and combining pairs of images. 
These DA methods tend to preserve the image semantics well but lack diversity and visual fidelity. 
A more promising way is to utilize generative models for DA. 
Recently, Diffusion Models (DM)~\cite{rombach2022high} have been introduced to generate highly photo-realistic synthetic image~\cite{azizi2023synthetic,he2022synthetic}.  
Despite the promise, this line of research faces two critical challenges. 
First, by focusing on the conditioning mechanism through the creation of diverse prompts and optimization of conditional embeddings, 
these methods 
are complex in design, potentially hindering practical applications.
Second, how to generate diverse samples through perturbing the diffusion process remains an open question. Naive approaches that perturb the diffusion latent (i.e., output of the denoising U-Net) 
during reverse diffusion yields limited data diversity. This is because diffusion latent has limited ability in defining image layout, as the \textit{cross-attention maps} between the denoising U-Net and the prompt tokens predominantly encode information related to \textit{shape} and \textit{structure}~\cite{hertz2023prompt}.
Moreover, the full reverse diffusion process is under-explored, which makes the trade-off between fidelity and diversity sub-optimal.


\begin{figure}[t]
\centering
\includegraphics[width=0.8\columnwidth]{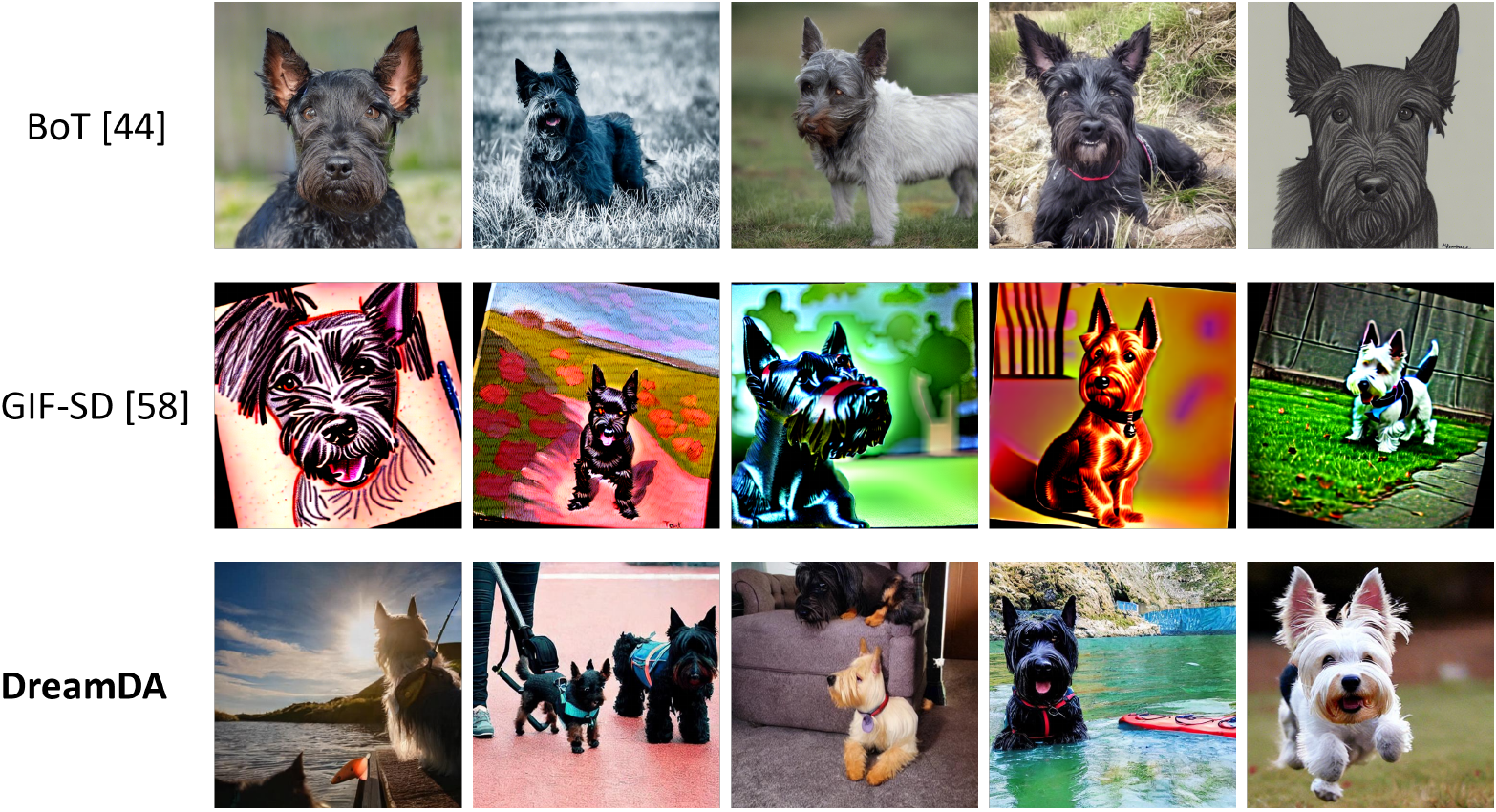}
\caption{Synthetic samples generated for "Scottish Terrier" from different diffusion-based DA methods.}
\label{fig1_DA_samples}
\vspace{-7mm}
\end{figure}

To this end, we propose a novel perturbation approach that enables the generation of photo-realistic in-distribution data for image classification tasks through the `lens' of diffusion models. 
The key idea is to generate highly diverse images that conform to the original data distribution by considering original training images as seeds and perturbing every step of the seeds' reverse diffusion processes.
We experiment with a variety of perturbation strategies and discover that adding Gaussian noise to the U-Net bottleneck layer of a diffusion model is an effective approach, achieving an excellent trade-off between image fidelity and diversity, as illustrated in Fig.~\ref{fig1_DA_samples}.

However, the diversity introduced implies that semantic labels of generated images are not guaranteed to align with those of the seed images. 
To address this issue, we introduce a new step-wise self-training paradigm Asymmetric Multi-Head Self-Training~(AMST) to train the classifier using our synthetic data. 
We introduce four classifiers to enforce consistency and confidence constraints such that predicted pseudo labels are ensured to be both accurate and reliable. 
We do not throw away samples that do not satisfy the constraints but treat them as unlabeled data and leverage a consistency regularization loss that encourages the same model prediction for similar data points. 
This helps the model learn general and perturbation invariant features.

We extensively evaluate the proposed DreamDA using five datasets that comprise four tasks by training standard backbones (e.g. Resnet50~\cite{he2016deep}) from scratch and the ImageNet1K~\cite{russakovsky2015imagenet} pre-trained checkpoint using a combined set of real and synthetic data. 
Notably, we compare DreamDA to widely-used DA techniques and recent diffusion model-based DA methods, and demonstrate consistent gains in accuracy for all datasets.
When applied to natural image datasets, synthetic images generated by DreamDA boost model accuracy by over 41\% and 4\% compared to the original dataset for models trained from scratch and pre-trained models, respectively.
The contributions are summarized as follows:
\begin{itemize}
    \item We propose a novel data augmentation framework named DreamDA that exploits knowledge in pre-trained diffusion models to generate diverse images that adhere to a real data distribution. 
    \item Technically, we propose to perturb the diffusion process by adding Gaussian noise to the U-Net inside diffusion models at each step during reverse diffusion. Moreover, we introduce a novel self-training method to tackle label inconsistency of synthesized data.
    \item DreamDA shows consistent improvements over representative DA techniques and diffusion model-based generative DA baselines across four tasks and five datasets. DreamDA outperforms the strongest diffusion-based DA baseline by over 7.4\% when trained from scratch on natural image datasets.
\end{itemize}

\section{Background}
\subsection{Diffusion Model}
Diffusion models (DM) are a family of probabilistic generative models that progressively destruct data by injecting noise in a forward process, then learn to reverse this process for sample generation~\cite{yang2022diffusion}. Specifically, the forward diffusion process gradually convert data samples \(x_0 \sim p_{data}(x)\) to a prior noise distribution with intermediate noisy data \(x_1, ..., x_T\) and \(x_T \sim \mathcal{N}(0,\pmb{I})\) by applying Gaussian transitions:
\(
q(x_t|x_{t-1}) = N(x_t;\sqrt{1-\beta_t}x_{t-1}, \beta_t\pmb{I}),
\)
where \(t \in \{1,...,T\}\) denote the timestep and \(\beta_t \in (0,1)\) is a predefined noise variance hyperparameter. Conveniently, \cite{sohl2015deep} observed that we can sample any arbitrary noisy data \(x_{t}\) from \(x_0\) with \(
x_t = \sqrt{\alpha_t}x_0 + \sqrt{1-\alpha_t}\epsilon,
\)
where \(\alpha_t \vcentcolon= \prod_{s=1}^t(1-\beta_s)\) and \(\epsilon \sim \mathcal{N}(0,\pmb{I}) \). When \(\alpha_T \approx 0\), \(x_T\) approximates the Gaussian \(\mathcal{N}(0,\pmb{I})\).

To generate samples, diffusion models learn the reverse diffusion process 
\(
p_{\theta}(x_{t-1}|x_t) = \mathcal{N}(x_{t-1};\mu_{\theta}(x_t, t),\Sigma_{\theta}(x_t,t) ),
\) where mean \(\mu_{\theta}(x_t, t)\) and variance \(\Sigma_{\theta}(x_t, t)\) are parameterized by neural networks. In practice, the variance \(\Sigma_{\theta}\) can be predefined or learned, and the mean \(\mu_{\theta}(x_t, t)\) can be parameterized using a noise predictor \(\epsilon_{\theta}\)~\cite{ho2020denoising,nichol2021improved} as \(\mu_{\theta}(x_t, t) = 1/{\sqrt{1-\beta_t}}(x_t- \beta_t/{\sqrt{1-\alpha_t}}\epsilon_{\theta}(x_t, t))\). The corrupted \(x_t\) are then iteratively denoised by subtracting the noise predicted by \(\epsilon_{\theta}\) with 
\(x_{t-1} = 1/{\sqrt{1-\beta_t}}(x_t- \beta_t/{\sqrt{1-\alpha_t}}\epsilon_{\theta}(x_t, t)) + \sigma_tz\),
where \(\sigma^2_t\) is the predefined or estimated variance of the reverse diffusion process, and \(z \sim \mathcal{N}(0,\pmb{I}) \). Note \(\sigma^2_t = \beta_t\) in Ho et al.~\cite{ho2020denoising}.


The noise predictor \(\epsilon_{\theta}\) is optimized by a sum of denoising score matching losses over timesteps \(t>0\), which is defined as 
\begin{equation} 
    L_t = \mathbb{E}_{x_0, \epsilon}[\frac{\beta_t^2}{2\sigma_t^2(1-\beta_t)(1-\alpha_t)} \norm{\epsilon - \epsilon_{\theta}(x_t,t)}^2]
\end{equation}
The total loss is therefore \(L_{vlb} = \sum^{T-1}_{t=1}L_t\) and is a variational lower bound~\cite{ho2020denoising}. Ho el at.~\cite{ho2020denoising} further propose to a simplified reweighting of \(L_t\) terms for improving sample quality with
\begin{equation}
\label{Lsimple}
L_{simple} = \sum^{T-1}_{t=1} \mathbb{E}_{x_0, \epsilon}[\norm{\epsilon - \epsilon_{\theta}(x_t,t)}^2] = \sum^{T-1}_{t=1}\lambda_tL_t
\end{equation}. 

We leverage the SOTA text-to-image (T2I) diffusion models as prior models and propose a novel approach to synthesize diverse images that mitigate the domain gap between synthetic and real data.

\subsection{Inversion} 
A common approach to image editing is inverting images to the latent space of generative models~\cite{mokady2023null}. For diffusion models, inversion can be achieved by performing forward diffusion using a pre-trained diffusion model to map \(x_0\) to \(x_T\). The inverted latent \(x_T\) can be used to initialize the reverse diffusion process to reconstruct the original image \(x_0\). Hence, image editing and manipulation are possible by conditioning the reverse diffusion process with desired editing conditions. 



The simplest inversion technique is DDIM inversion~\cite{song2020denoising,dhariwal2021diffusion}, which assumes that an ODE reverse diffusion process is reversible. The ODE reverse diffusion is defined as:
\begin{equation}
    \resizebox{.8\hsize}{!}{$x_{t-1} =\sqrt{\alpha_{t-1}} \underbrace{ (\frac{x_t-\sqrt{1-\alpha_{t}}\epsilon_{\theta}(x_t,t)}{\sqrt{\alpha_t}})}_{\text{predicted} \ x_0} + \underbrace{\sqrt{1-\alpha_{t-1}}\epsilon_{\theta}(x_t,t)}_{\text{direction pointing to} \ x_t}$}
\label{DDIM_sampling}
\end{equation}
For simplicity, Equation.~\ref{DDIM_sampling} can be rewritten as 
\begin{equation}
    x_{t-1} =S_t(\epsilon_{\theta}(x_t,t)) + D_t(\epsilon_{\theta}(x_t,t)),
\label{DDIM_sampling2}
\end{equation}
where \(S_t(\epsilon_{\theta}(x_t,t))\) and \(D_t(\epsilon_{\theta}(x_t,t))\) denote the predicted \(x_0\) and direction pointing to \(x_t\), respectively.

In the limit of small steps, DDIM inversion (i.e. forward diffusion process) using a pre-trained diffusion model is therefore:
\begin{equation}
    \resizebox{.8\hsize}{!}{$x_{t+1} =\sqrt{\alpha_{t+1}} (\frac{x_t-\sqrt{1-\alpha_{t}}\epsilon_{\theta}(x_t,t)}{\sqrt{\alpha_t}}) + \sqrt{1-\alpha_{t+1}}\epsilon_{\theta}(x_t,t)$}
\end{equation}

\section{Related Works}
\noindent\textbf{Text-to-image models.} 
Recent T2I models including GLIDE~\cite{nichol2021glide}, Imagen~\cite{saharia2022photorealistic}, Stable Diffusion (SD)~\cite{rombach2022high}, Dall-E~\cite{ramesh2021zero}, and Muse~\cite{chang2023muse} have the capacity to produce realistic images with high fidelity from text conditions. In this paper, we leverage SD~\cite{rombach2022high} as a prior model and propose a novel method to synthesize images with high diversity. 


\noindent\textbf{Image Generation for Classification.} 
In computer vision, generative DA has been well studied to provide additional data for training predictive models~\cite{li2022bigdatasetgan, he2022synthetic,jahanian2021generative,azizi2023synthetic}. DatasetGANs~\cite{zhang2021datasetgan,li2022bigdatasetgan} explored Generative Adversarial Networks for data synthesis, but suffer from a lack of diversity.

Recently, diffusion models have been leveraged for DA. He et al.~\cite{he2022synthetic} employed GLIDE as a prior model for DA, enhancing performance in data-limited scenarios.
Azizi et al~\cite{azizi2023synthetic} fine-tuned Imagen~\cite{saharia2022photorealistic} to mitigate the domain gap between generated and real data. GIF-SD~\cite{zhang2022expanding} perturbs the image latent with noise and optimizes the noisy latent with a diversity loss. DI~\cite{zhou2023training} performs textual inversion~\cite{gal2022image} and perturbs the conditioning embedding. BoT~\cite{shipard2023diversity} proposes a series of techniques to improve the diversity of text prompts and conditioning strengths. Several other works focus on designing or optimizing prompts for SD~\cite{dunlap2023diversify, bansal2023leaving, yuan2022not, sariyildiz2023fake, shin2023fill,shivashankar2023semantic,trabucco2023effective} and demonstrate promising results. Other studies have also explored data synthesis for semantic segmentation and object detection tasks~\cite{wu2023diffumask,wu2023datasetdm}. However, these works do not fully exploit the diffusion process for diverse sample generation. In contrast, we propose an inversion-based latent perturbation method within the reverse diffusion process, effectively promoting the generation of diverse samples that adhere to the real data distribution.

\begin{figure*}[t]
\centering
\includegraphics[width=\textwidth]{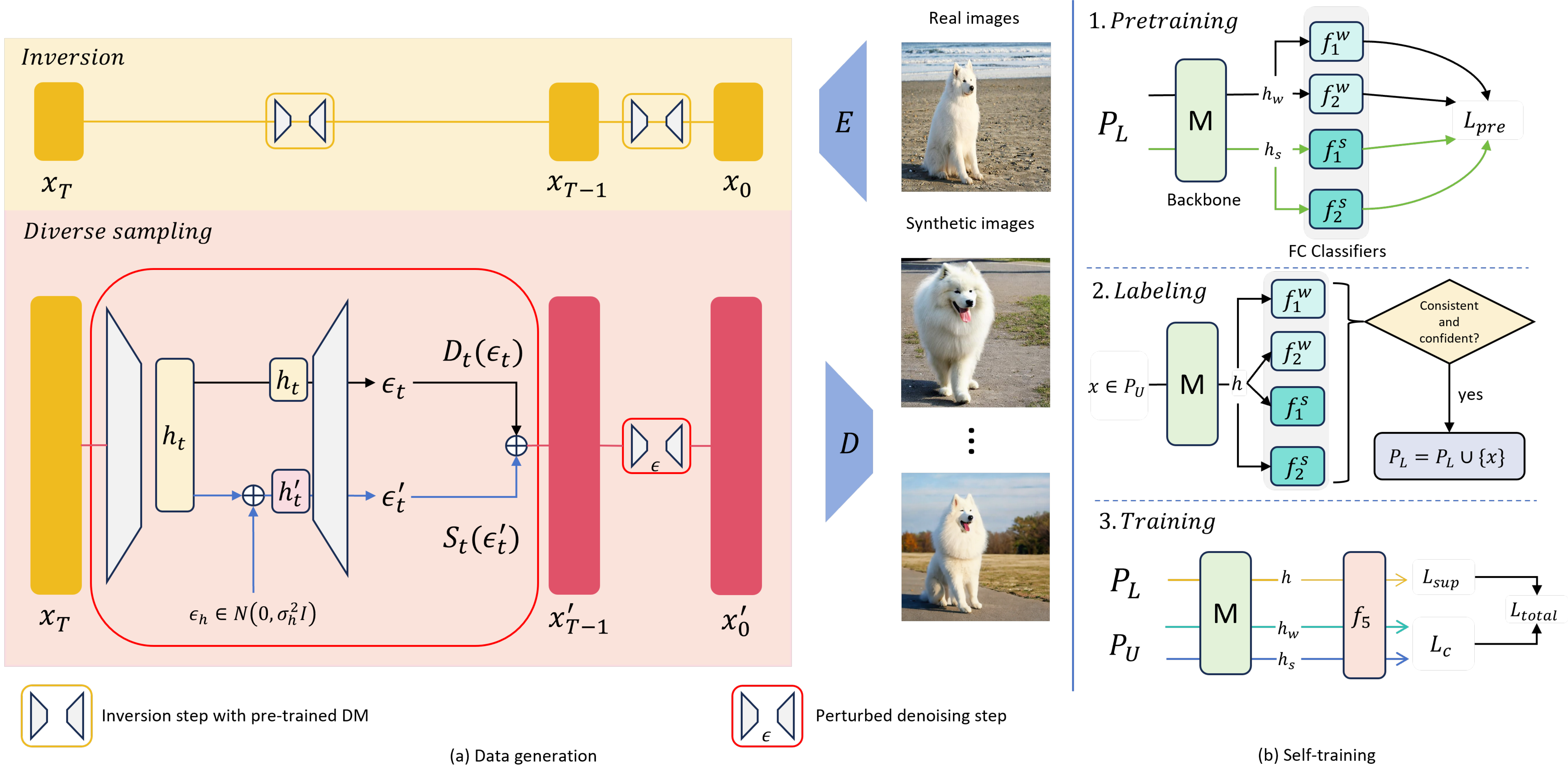}
\caption{Our proposed generation and self-training framework. Here, \(\mathbf{E}\) and \(\mathbf{D}\) represent the VAE encoder and decoder for SD, respectively. (a) illustrate latent perturbation for diverse sample generation. We invert image latents \(x_0\) to \(x_T\) using SD. 
Subsequently, we perturb each step of the reverse diffusion process by adding Gaussian noise to the U-Net bottleneck feature of the diffusion model, yielding diverse variations of the real images. (b) illustrate our training approach using synthesized data. We train four auxiliary classifiers to make accurate and reliable class label predictions for synthesized images to train the fifth main classifier.}
\label{method_fig}
\vspace{-5mm}
\end{figure*}

\section{Methodology}

\subsection{Overview} 
The overview of DreamDA is illustrated in Fig.~\ref{method_fig}. 
The goal of DreamDA is to synthesize data that complement the original real dataset with novel and informative samples, which adhere to the original data distribution while providing novel and diverse image content.
In order to mitigate the domain gap between synthetic and real data during image generation, we invert image latent \(x_0\) to \(x_T\) using SD. Next, we generate diverse samples by perturbing the U-Net bottleneck feature of the diffusion model for each step in the reverse diffusion process. In essence, we are progressively making random edits to the image, allowing the synthesis of diverse images that adhere to the original data distribution.
We note that the semantics of synthesized images may not align with the seed image. Therefore, to diminish errors due to incorrect class labels, we propose AMST, a unified framework for pseudo labeling and training. Here, four auxiliary classifiers are used to ensure accurate and reliable labels, while consistency regularization is employed to learn perturbation invariant features.

\subsection{Latent Perturbation}
In this part, we aim to improve the diversity of synthetic data.
Our main idea is to treat each individual sample within the real dataset as a \emph{seed} image, and subsequently utilize a diffusion model to generate informative variations of each seed image that contain novel and distinctive content. 
It is important that the synthesized data preserves the original semantics and other essential characteristics of the real images as much as possible. 
To accomplish this, we resort to the inversion approach~\cite{song2020denoising} that maps a seed image to a noisy latent representation, effectively capturing the semantics and content of the seed image. 
However, it is unclear how to generate diverse samples given an image inversion trajectory. 
Thus, we conducted a preliminary study and found that adding noise to the U-Net bottleneck layer of SD at each step during the reverse diffusion process can yield diverse samples.


\noindent\textbf{Preliminary Study.}
Given an inversion trajectory, the original seed image can be reconstructed by running reverse diffusion. We seek a perturbation method during reverse diffusion such that the resulting synthetic image and the seed image share similar semantics but the synthetic image contains novel content. 
Our key insight is that the reverse diffusion process governs the entire generation procedure, so perturbing the reverse diffusion process can generate diverse variations of the seed image. A natural question arises in how to perturb the reverse diffusion process. We conducted a preliminary study to investigate various perturbation methods.

The standard DDIM inversion~\cite{song2020denoising,dhariwal2021diffusion} was adopted to invert image latents \(x_0\) to its prior noise distribution \(x_T\). Next, we study two types of methods to perturb the diffusion process: (1) directly adding noise to the latent \(x_t\) at different timesteps and (2) adding noise to the features inside the diffusion model \(\epsilon_{\theta}\), altering the noise prediction for "predicted \(x_0\)" \(S_t\) while keeping "direction pointing to \(x_t\)" \(D_t\) unchanged in Equation.~\ref{DDIM_sampling2}.

For perturbing latent \(x_t\), we experimented with adding Gaussian noise to the start \(x_T\), intermediate steps \(x_t, t\in \{1,...,T-1\}\), and end \(\tilde{x}_0\) of the reverse diffusion process. To investigate perturbation inside the diffusion model, we tried adding Gaussian noise to the feature map of each layer of the U-Net encoder, bottleneck layer, and decoder for each step during reverse diffusion. We evaluate generated data using the Oxford-IIT Pet~\cite{parkhi2012cats} dataset. Specifically, a standard Resnet50~\cite{he2016deep} classifier was trained from scratch using a combined set of synthetic and real data. We ran each perturbation method three times using a noise scale of \{0.1, 1, 1.5\} and reported the highest accuracy. 

\begin{wraptable}{r}{7cm}
\vspace{-0.5cm}
\centering
\scalebox{0.9}{
\begin{tabular}{l|cc} 

\toprule

& Perturbation location  & Acc \\
\midrule
\midrule
\multirow{3}{*}{Diffusion latent} & Latent \(\tilde{x}_0\)& 56.9 \\
&Latent \(x_T\)& 59.3  \\
&Continuous latent \(\{x_t\}^{T-1}_1\) & 60.6  \\
\midrule
\multirow{3}{*}{Model latent} &U-Net encoder & 44.2 \\
&U-Net decoder & 67.3 \\
&U-Net bottleneck & 73.4 \\

\bottomrule
\end{tabular}
}
\caption{Investigation on various perturbation methods given an inversion trajectory. We generate ten times more images than the original dataset and train Resnet50 from scratch using a combined set of synthetic and real images. We report the highest accuracy over noise scales \{0.1, 1, 1.5\} on the Oxford-IIT Pet~\cite{parkhi2012cats} dataset.\vspace{-4mm}}
\label{preliminary_study}
\end{wraptable}

The results presented in Table~\ref{preliminary_study} provide empirical evidence supporting the effectiveness of adding noise to the U-Net bottleneck layer. This approach achieves an excellent trade-off between image fidelity and diversity. Unsurprisingly, we observe limited diversity resulting from methods introducing perturbations to the diffusion latent \(x_t\). This limitation arises from the fact that the overall image layout cannot be substantially altered through diffusion latent perturbations, as the \textit{cross-attention maps} between the denoising U-Net and the prompt embedding encapsulate the majority of information related to \textit{shape} and \textit{structure}~\cite{hertz2023prompt}. In contrast, bottleneck layer perturbations through noise injection allow adjustments to the image layout during \textit{cross-attention} in the U-Net decoder.
Furthermore, the U-Net bottleneck feature serves as a compact latent representation~\cite{DBLP:journals/corr/abs-1810-10331} itself for the corresponding diffusion latent, meaning that local perturbations in the U-Net bottleneck layer can achieve larger-scale variations in the generated final image.

\noindent\textbf{Proposed Perturbation Solution.} 
On the basis of our preliminary study, we propose to perturb the reverse diffusion process by perturbing the U-Net bottleneck feature \(h_t\) of diffusion models at every step \(t\) during reverse diffusion initialized with an inverted latent \(x_T\). Precisely, we add Gaussian noise \(\epsilon_h\) to obtain a noisy bottleneck feature \(h'_t= h_t + \epsilon_h\). Using \(h'_t\), we reformulate Equation.~\ref{DDIM_sampling2} with a perturbed "prediction of \(x_0\)" as
\begin{equation}
    x'_{t-1} =S_t(\tilde{\epsilon}_{\theta}(x'_t,t,\epsilon_h)) + D_t(\epsilon_{\theta}(x'_t,t)),
    \label{eq_perturbation}
\end{equation}
where \(x'_T = x_T\), \(\tilde{\epsilon}_{\theta}(x'_t,t,\epsilon_h)\) is the resulting output of the noise predictor \(\epsilon_{\theta}\) obtained by passing the perturbed bottleneck feature \(h'_t\) to the U-Net decoder, and \(\epsilon_h \sim \mathcal{N}(0,\sigma_h^2I)\) with \(\sigma_h^2\) as a noise scale hyperparameter. Intuitively, \(h'_t\) randomly and iteratively shifts the prediction of the noise required to reach \(x_0\), which yields diverse variations in the generated synthetic images.

In practice, we randomly generate Gaussian noise \(\epsilon_h\) and add it to \(h_t\) on the fly. The result is used as the input to the U-Net decoder at  step $t$ in the reverse diffusion process. To generate a set of \(n\) diverse samples originating from a given seed image \(x_0\), we perform inversion once using \(x_0\), and run \(n\) distinct reverse diffusion processes subsequently.



\subsection{Asymmetric Multi-Head Self-Training}

Our diffusion perturbation approach is random,
which implies that synthesized samples are not guaranteed to have consistent labels with the seed images. 
To mitigate the influence of incorrect labels,
we propose Asymmetric Multi-head Self-Training~(AMST) to effectively utilize synthesized data for classification tasks. 
As shown in Figure~\ref{method_fig}~(b), 
AMST consists of a backbone with five classifiers. 
The four auxiliary classifiers are responsible for generating pseudo labels, while the main classifier is trained on both real and synthesized data. 
AMST has three stages, namely, pre-training with real data, pseudo-labeling synthetic data, and training with labeled and unlabeled data. 

\noindent\textbf{Stage 1:} 
We augment each input image twice, producing a weakly augmented version \(x_w\) and a strongly augmented version \(x_s\). As illustrated in Fig.~\ref{method_fig} (b), the first pair of auxiliary classifiers take \(x_w\) as input, while the other two auxiliary classifiers take \(x_s\) as input. To encourage each pair of auxiliary classifiers to learn distinct features from the same input, we impose an orthogonal constraint over their weights as an additional loss function. 
The complete pre-training loss is defined as
\begin{equation}
\small
\begin{split}
    L_{pre} & = \frac{1}{n}\sum^n_{i=1}[\mathcal{H}(f^{w}_1(h^{w}_i), y_i) + \mathcal{H}(f^{w}_2(h^{w}_i), y_i) \\
    &+ \mathcal{H}(f^{s}_1(h^{s}_i), y_i) + \mathcal{H} (f^{s}_2(h^{s}_i), y_i)] 
    + \lambda(|{W^{w}_1}^TW^{w}_2| +  |{W^{s}_1}^TW^{s}_2|),
\end{split}
\end{equation}
where \(h^w_i\) and \(h^s_i\) are the features of sample \(i\) extracted by the backbone for \(x_w\) and \(x_s\), respectively, \(\{f^{w}_1,f^{w}_2,f^{s}_1,f^{s}_2\}\) and \(\{W^{w}_1,W^{w}_2,W^{s}_1,W^{s}_2\}\) denote the four auxiliary classifiers and their weight matrices, respectively, \(
y_i\) is the ground-truth label, \(\mathcal{H}\) is the standard cross entropy loss,  and \(\lambda\) is a weighting hyperparameter, which is set to 0.001 in all our experiments.

\noindent\textbf{Stage 2:} We use the trained auxiliary classifiers to predict pseudo-labels for our generated images. 
We assume a pseudo label is reliable if all four classifiers are consistent and confident. 
Being consistent means all auxiliary classifiers predict the same class label, whereas confidence is measured with an uncertainty score \(u_i\). 
We estimate \(u_i\) using MC Dropout, which passes the same image into a dropout-enabled model and computes the variance of model output~\cite{gal2016dropout}. 
The confidence constraint is defined as \(\max_i(u_i) < \tau\), where \(\tau\) is a threshold set to 0.01. 
We predict labels for each synthesized image and check if the labels satisfy the consistency and confidence constraints. We initialize a labeled set \(P_L\) and unlabeled set \(P_U\) with the real and synthetic data, respectively. For each synthetic image, if the consistency and confidence constraints are satisfied, its predicted label is kept and the image is removed from the unlabeled set \(P_U\) and appended to the labeled set \(P_L\).

\noindent\textbf{Stage 3:} We train the main classifier \(f_5\) using a 
combination of losses for the labeled and unlabeled sets. Specifically, the cross entropy loss for the labeled set is
\begin{equation}
    L_{sup} = \sum_{x_i,y_i) \in P_L}\mathcal{H}(f_5(h_i), y_i),
\end{equation}
where \(h_i\) denotes the original image feature extracted by the backbone. For the unlabeled set, the consistency regularization loss  is
\begin{equation}
    L_{c} = \sum_{x_i \in P_U}\lambda(y^w_i)\mathcal{H}(y^w_i), y^s_i),
\end{equation}
where \(y^w_i = f_5(h^{w}_i)\) and \(y^s_i = f_5(h^{s}_i)\) denote the predictions for the weakly and strongly augmented input, respectively, and  \(\lambda(\cdot)\) is a weighting function defined according to~\cite{chen2023softmatch}. The total loss for training \(f_5\) is therefore \(L_{total} = L_{sup} + L_{c}\).

Notably, AMST shares similarities with Tri-training~\cite{saito2017asymmetric}, which focuses on pseudo-labeling for domain adaptation. However, there exist key differences between AMST and Tri-training. AMST extends Tri-training to a unified framework that bridges the gap between the label predictor and image generator by ways of augmentation, orthogonal prediction, and consistency regularization. Our AMST ensures prediction accuracy for both real (source) and synthetic (target) data. In contrast, Tri-training may forget the source knowledge as the supervision signal is only from pseudo labels. In practice, AMST is a crucial component in our pipeline, giving rise to substantial gains in performance as demonstrated in Table \ref{ablation_overall_amst_noise} and Table \ref{ablation_AMST}.

\newcommand{\cmark}{\ding{51}}%
\newcommand{\xmark}{\ding{55}}%

\section{Experiment}


\subsection{Setup}
\noindent\textbf{Datasets.} We conduct a comprehensive evaluation of our approach across a diverse set of five datasets. These datasets consist of various data types, including natural and medical images. 
For natural image datasets, we employ the widely used Caltech101~\cite{fei2004learning}, Stanford Cars (Cars)~\cite{krause2013collecting}, and Oxford-IIIT Pet (Pets)~\cite{parkhi2012cats}. 
We extend our analysis to include a chest X-ray medical dataset: Shenzhen Tuberculosis (Shenzhen TB)~\cite{jaeger2014two}, which aims to classify whether a chest X-ray image shows manifestations of tuberculosis. Additionally, we use STL-10~\cite{coates2011analysis} for semi-supervised learning experiments. A detailed overview of the dataset statistics is presented in Table~\ref{datasetStats}.


\definecolor{ForestGreen}{RGB}{34,139,34}

\begin{table}[t]
\small
\centering
\setlength{\abovecaptionskip}{0.3cm}
\setlength{\belowcaptionskip}{-0.5cm}
\setlength\tabcolsep{2pt}
\scalebox{0.75}{
\begin{tabular}{llccc} 
\toprule
Datasets & Type of classification & \# Train samples & \# Classes & \# Samples per class \\
\hline
\hline
Caltech101~\cite{fei2004learning} & Coarse-grained & 3030 & 101 & 30 \\
Oxford-IIT Pet~\cite{parkhi2012cats} & Fine-grained & 3680 & 37 & 99 \\
Stanford Cars~\cite{krause2013collecting} & Fine-grained & 8144 & 196 & 42 \\
STL-10~\cite{coates2011analysis} & SSL & 5000 & 10 & 500 \\
Shenzhen Tuberculosis~\cite{jaeger2014two} & Medical & 463 & 2 & 232 \\
\bottomrule
\end{tabular}}
\caption{Dataset statistics. Note that STL-10 has 100,000 unlabeled images in addition to labeled ones.}
\label{datasetStats}
\end{table}

\noindent\textbf{Evaluation.} Two types of metrics are used to conduct performance evaluation: classification accuracy and distribution similarity. To evaluate classification accuracy, we train standard deep learning backbones using a dataset comprising both synthetic and real images. During the training process, we utilize standard augmentations, namely random cropping and horizontal flipping. 
To measure the similarity between real and synthetic data, we use the Fréchet Inception Distance~\cite{heusel2017gans} (FID) and Maximum Mean Discrepancy~\cite{tolstikhin2016minimax} (MMD). 





\noindent\textbf{Implementation.}
We use Stable Diffusion v1.5 from Huggingface and perform weighted fine-tuning using the original training data with hyperparameters set according to~\cite{choi2022perception}. The fine-tuned SD is used in our pipeline for further experiments. We leverage template prompts of the form "A photo of a \{class\}" in preliminary studies, whereas GPT-3.5-turbo is asked to provide diverse prompts for all other DreamDA experiments\footnote{For example, the following prompt was given to GPT-3.5-turbo to generate prompts for the "Persian" class in the Oxford-IIIT Pet dataset: 'Please provide 100 language descriptions for random scenes that contain only the class "Persian" from the Oxford-IIIT Pet dataset. Each description should be different and contain a minimum of 15 words. These descriptions will serve as a guide for Stable Diffusion in generating images.'}.

For latent perturbation, we utilize CycleDiffusion~\cite{wu2023latent} as the inversion method and use isotropic Gaussian noise with \(\sigma_h =3\) for perturbation. 

\begin{table}[t]
\centering
\setlength{\abovecaptionskip}{0.3cm}
\setlength{\belowcaptionskip}{-0.5cm}
\scalebox{0.88}{
\begin{tabular}{c|lccc|c|c} 

\toprule
\multirow{2}{*}{Regime} & \multirow{2}{*}{Method}& \multicolumn{3}{c|}{Natural Images}  & Medical Images & SSL \\
&  & Caltech101 & Cars & Pets\ \  & Shenzhen TB  & STL-10 \\

\midrule
\midrule
\multirow{8}{*}{Pretrained} &Original & 89.1 & 89.2  & 86.9   & 88.3  &72.4 \\
&CutMix~\cite{yun2019cutmix} & 88.9 & 89.1  & 87.0 & 89.1 & 72.4 \\
&Random Erase~\cite{zhong2020random} & 89.2 & 89.3  & 87.1  & 89.5   & 72.5   \\
&RandAugment~\cite{cubuk2020randaugment} & 89.5 & 89.7   & 87.1  & 89.7  & 72.7   \\
&Stable Diffusion~\cite{rombach2022high} & 87.2 & 92.8  & 87.3  & 90.4  &72.8 \\
& SDEdit~\cite{meng2021sdedit}  &88.2 &91.4 &88.3 & 90.2  & 72.9 \\
& DI~\cite{zhou2023training} & 87.8 & 92.3  & 88.4 & 89.5 & 72.7\\
& BoT~\cite{shipard2023diversity} & 88.4 & 92.5   & 88.6 & 88.7 & 73.0 \\
& GIF-SD~\cite{zhang2022expanding} & 86.4 & 93.1  & 87.5 & 89.2 & 72.6\\
\rowcolor[rgb]{0.867, 0.922, 0.969} &DreamDA  & \textbf{93.1}{\scriptsize({\color{ForestGreen}+3.6})} & \textbf{94.5}{\scriptsize({\color{ForestGreen}+1.4})} & \textbf{93.2}{\scriptsize({\color{ForestGreen}+4.6})} & \textbf{91.1}{\scriptsize({\color{ForestGreen}+0.7})} & \textbf{74.1}{\scriptsize({\color{ForestGreen}+1.1})}\\

\midrule
\midrule
\multirow{8}{*}{Scratch} & Original & 43.9 & 46.1   &  43.8& 72.7  & 25.1 \\
& CutMix~\cite{yun2019cutmix} & 55.8 & 60.1   & 58.6& 74.2 & 27.1\\
& Random Erase~\cite{zhong2020random} & 55.3 & 62.3  & 55.3& 74.1  & 26.9   \\
& RandAugment~\cite{cubuk2020randaugment} & 60.1 & 65.8   & 61.5 &  75.5 & 27.6    \\
& Stable Diffusion~\cite{rombach2022high} & 69.6 & 71.9  & 70.1& 76.5 & 28.7 \\
& SDEdit~\cite{meng2021sdedit}  &76.2 &68.4  &72.6 & 78.6  & 29.9 \\
& DI~\cite{zhou2023training} &73.1 & 74.8 & 74.8 & 79.9 & 28.1 \\
& BoT~\cite{shipard2023diversity} & 77.9 & 76.2  & 73.7 & 75.8 & 29.4 \\
&GIF-SD~\cite{zhang2022expanding} & 65.1 & 75.7  & 73.4 & 77.3   & 27.5\\
\rowcolor[rgb]{0.867, 0.922, 0.969} &DreamDA  & \textbf{85.3}{\scriptsize({\color{ForestGreen}+7.4})} & \textbf{87.1}{\scriptsize({\color{ForestGreen}+10.9})} & \textbf{86.5}{\scriptsize({\color{ForestGreen}+11.7})} & \textbf{83.5}{\scriptsize({\color{ForestGreen}+3.6})} & \textbf{34.6}{\scriptsize({\color{ForestGreen}+4.7})} \\



\bottomrule
\end{tabular}}
\caption{Top1 accuracy of Resnet50 trained from scratch and ImageNet pre-trained checkpoint on various datasets. Notably, the datasets include both coarse-grained classification datasets such as Caltech101, and fine-grained ones such as Pets and Cars. Shenzhen TB represents a medical dataset, while STL-10 is a semi-supervised learning dataset.
For a fair comparison to~\cite{zhang2022expanding}, the Pets dataset was expanded by a scale of 30, while Cars and Caltech101 were expanded by 20 times. All other datasets were expanded by a scale of 10. We report the average accuracy of three runs with different random seeds.}
\label{main}
\end{table}

\subsection{Baselines} 
We compare DreamDA with conventional and diffusion-based data augmentation (DA) methods. Specifically, the following three representative DA techniques were evaluated: CutMix~\cite{yun2019cutmix}, RandAugment~\cite{cubuk2020randaugment}, and Random Erase~\cite{zhong2020random}. Among diffusion-based baselines, we compare with Stable Diffusion~\cite{rombach2022high}, noise perturbation approaches (GIF-SD~\cite{zhang2022expanding}, DI~\cite{zhou2023training}), prompt engineering/optimization methods (BoT~\cite{shipard2023diversity}, DI~\cite{zhou2023training}), and SDEdit~\cite{meng2021sdedit} (a representative image translation method). 
The performance of the original real datasets is denoted as \textit{Original}. For SSL, we follow SoftMatch~\cite{chen2023softmatch} to train models on the original dataset, and expand the original labeled set by a scale of 10 using various DA approaches while the original unlabeled set remains unmodified.


\subsection{Main Results} 
We conduct evaluations on five widely used datasets across four tasks, namely coarse-grained classification, fine-grained classification, medical image classification, and semi-supervised learning. 
In addition, we demonstrate robustness by evaluating our method on multiple modern deep learning architectures. 
Through these experiments, we provide compelling evidence of the efficacy and wide-ranging applicability of our proposed method in diverse practical scenarios. The results are reported in Table~\ref{main}.

\noindent\textbf{Fully-Supervised Learning.} 
As can be seen, DreamDA consistently outperforms other DA and diffusion-based baselines across natural and medical image classification tasks. 
For the natural image datasets, DreamDA achieves over
41\% and 4\% increase in absolute accuracy compared to the original dataset, for the scratch and pretrained training regimes, respectively. Moreover, DreamDA outperforms the strongest diffusion-based DA baselines on natural image datasets by over 7.4\% and 1.4\% for the scratch and pre-trained training approaches, respectively.
In the case of the medical dataset, our approach continues to exhibit superior performance compared to diffusion and DA baselines, surpassing the strongest diffusion-based baseline by 3.6\% when trained from scratch on the Shenzhen TB datasets. These compelling results serve as empirical evidence that DreamDA establishes a new state-of-the-art performance in the realm of generative data augmentation.

\noindent\textbf{Semi-Supervised Learning.} 
To further demonstrate the generalizability of our approach, we evaluate our approach on STL-10, which is a common semi-supervised learning (SSL) dataset. Following Chen et al.~\cite{chen2023softmatch}, we randomly sample 40 labels for training. For DreamDA, we use latent perturbation and asymmetric multi-head labeling to generate labeled images. Models in all SSL experiments are trained using SoftMatch~\cite{chen2023softmatch} with the original unlabeled set unchanged. Table~\ref{main} shows that our approach yields consistent accuracy gains compared to DA and diffusion-based baselines, exhibiting a significant 9.5\% improvement over the original dataset when trained from scratch. This confirms that our method can generate diverse images with accurate labels.

\definecolor{dreamda}{rgb}{0.867, 0.922, 0.969}

\newcolumntype{Y}{>{\centering\arraybackslash}X}
\begin{table}[t]

\
\centering

\setlength{\abovecaptionskip}{0.3cm}
\setlength{\belowcaptionskip}{-0.5cm}
\scalebox{0.85}{
\begin{tabularx}{\textwidth}{lYYYYYcc}

\toprule

Model  & Original & SD & SDEdit & DI & BoT & GIF-SD & DreamDA \\
\midrule
\midrule
Resnet50 & 43.8 & 70.1 & 72.6 & 74.8 & 73.7 & 73.4 & \textbf{86.5} \\
ResNeXt-50 & 44.6 & 73.4 & 74.1 & 75.0 & 77.8 & 75.3 & \textbf{86.9}\\
Swin Transformer-v2-tiny & 45.4 & 77.9 & 79.5 & 78.3 & 81.2 & 76.1 & \textbf{91.3} \\
MobilteNet-v2 & 39.7 & 68.1 & 68.9 & 72.9 & 71.4 & 66.8 & \textbf{82.5} \\
Avg& 43.4 & 72.4 & 73.8 & 75.4 & 76.0 & 72.9 & \textbf{86.8} \\

\bottomrule
\end{tabularx}}
\caption{Performance of various neural architectures trained from scratch using different dataset expansion techniques applied to the Oxford-IIT Pet dataset.}
\label{results_backbones}
\end{table}

\noindent\textbf{Versatility.} To verify the versatility of our approach, we train three additional representative deep learning backbones from scratch. Table~\ref{results_backbones} provides the performance of ResNeXt~\cite{xie2017aggregated}, Swin Transformer-v2~\cite{liu2022swin}, and MobileNet-v2~\cite{sandler2018mobilenetv2}. We observe that our proposed approach consistently outperforms diffusion-based baselines, 
proving the effectiveness of the expanded dataset in training various backbones.

\begin{table}[t]
\setlength{\abovecaptionskip}{0.3cm}
\setlength{\belowcaptionskip}{-0.1cm}
\centering
\setlength\tabcolsep{4pt}
\scalebox{1}{
\begin{tabular}{lccc} 

\toprule
\multirow{2}{*}{Method}& \multicolumn{3}{c}{Natural Images} \\
 & Caltech101 & Cars & Pets \\
\midrule
\midrule
\rowcolor{ggray} \multicolumn{4}{c}{Metric: FID $\downarrow$} \\
Stable Diffusion~\cite{rombach2022high} & 29.7 & 31.2 & 29.4  \\
SDEdit~\cite{meng2021sdedit}  & 20.3 & 20.9 &21.2 \\
DI~\cite{zhou2023training} & 11.0 & 12.3 & 13.8 \\
BoT~\cite{shipard2023diversity} & 31.3 & 33.8  & 30.7  \\
GIF-SD~\cite{zhang2022expanding} & 41.8 & 39.2  & 44.3  \\
\rowcolor[rgb]{0.867, 0.922, 0.969} DreamDA & \textbf{9.4} & \textbf{10.8} & \textbf{13.5} \\

\midrule
\midrule
\rowcolor{ggray} \multicolumn{4}{c}{Metric: MMD $\downarrow$} \\
Stable Diffusion~\cite{rombach2022high} & 0.065  & 0.069  &0.063   \\
SDEdit~\cite{meng2021sdedit}  & 0.062 & 0.065 & 0.062 \\
DI~\cite{zhou2023training} & 0.061 & 0.063  &  0.058 \\ 
BoT~\cite{shipard2023diversity} & 0.069 & 0.073 & 0.065 \\
GIF-SD~\cite{zhang2022expanding} & 0.083 & 0.086  & 0.079  \\
\rowcolor[rgb]{0.867, 0.922, 0.969} DreamDA & \textbf{0.060} & \textbf{0.061} & \textbf{0.056}  \\
\bottomrule
\end{tabular}}
\caption{Comparison of FID and MMD scores between DreamDA and other diffusion-based baselines.}
\label{distribution_similarity}
\renewcommand\thetable{7}
\end{table}

\noindent\textbf{Distribution Similarity.} 
Table~\ref{distribution_similarity} compares FID and MMD scores of synthetic images generated by diffusion-based DA.
We observe that DreamDA consistently outperforms the diffusion-based baselines in terms of FID and MMD scores, with particularly noteworthy improvements observed in the former. Together with the results presented in Table~\ref{main}, these findings offer compelling evidence of DreamDA's capacity to generate diverse in-distribution images.


\noindent\textbf{Computational Costs.} 
The average running times of SD, SDEdit, BoT, DI, GIF-SD, and DreamDA for generating one 512$\times$512 image using 8 H800 GPUs are 0.62s, 0.50s, 0.61s, 1.04s, 0.43s, and 0.56s, respectively. Thus, DreamDA achieves compelling performance without significant computational overhead. It is worth noting that the data generation procedure only has to be conducted once. The synthetic data can be saved and used repeatedly, eliminating the need to synthesize data on the fly every time a model is trained. 







\subsection{Ablation Studies}

\begin{table}[t]
\setlength{\abovecaptionskip}{0.3cm}
\setlength{\belowcaptionskip}{-0.1cm}
\centering
\scalebox{1.0}{
\begin{tabular}{lccc} 

\toprule

   & Latent Perturbation  &  AMST &  Accuracy \\
\hline
\hline
& & & 79.7 \\
& &\cmark & 80.9 \\
&\cmark & & 84.1 \\
 &\cmark &\cmark & 86.5 \\

\bottomrule
\end{tabular}}
\caption{Ablation on the overall effect of latent perturbation and AMST.\vspace{-2mm}}
\label{ablation_overall_amst_noise}
\end{table}
In order to evaluate the effectiveness of the proposed components, namely latent perturbation and AMST, ablation studies were conducted using the Pets~\cite{parkhi2012cats} dataset and the Resnet50~\cite{he2016deep} backbone trained from scratch. The results are presented in Table~\ref{ablation_overall_amst_noise}, which shows a significant decline of 5.6\% and 2.4\% in performance upon removal of latent perturbation and AMST, respectively. 
These results indicate that generating diverse in-distribution samples with latent perturbation and AMST are indeed beneficial for predictive models. Notably, DreamDA without latent perturbation and AMST correspond to SD with GPT prompts, which improves performance by 9.6\% compared to SD with template prompts (Table~\ref{main}). The inclusion of latent perturbation and AMST further substantially improves performance, highlighting the complementary nature between our proposed method and existing prompt engineering approaches. Therefore, the integration of more advanced prompt engineering techniques holds the potential for additional performance enhancements. Further investigation was carried out by individually ablating latent perturbation and AMST, as discussed below.

\begin{minipage}[t]{\columnwidth}
\begin{minipage}{0.35\columnwidth}
\centering
\vspace{0.2cm}
\makeatletter\def\@captype{table}
  \setlength{\abovecaptionskip}{0.1cm}
  \setlength{\belowcaptionskip}{0.2cm}
\scalebox{0.7}{
\begin{tabular}{lc} 

\toprule

Method  & Accuracy \\
\hline
\hline
Perturbation at \(x_0\) &  67.5 \\
Perturbation at \(\{x_t\}^{T-1}_1\) & 76.7  \\
Perturbation at U-Net decoder &  82.2  \\
DreamDA &  86.5  \\

\bottomrule
\end{tabular}}
\caption{Comparison among different perturbation techniques.}
\label{ablation_noise_inversion}
\end{minipage}
\quad
\begin{minipage}{0.55\columnwidth}
\centering
\vspace{0.4cm}
\makeatletter\def\@captype{table}
  \setlength{\abovecaptionskip}{0.1cm}
  \setlength{\belowcaptionskip}{0.2cm}
\scalebox{0.7}{
\begin{tabular}{lc} 

\toprule

Method  & Accuracy \\
\hline
\hline
Soft pseudo label &  84.6 \\
CLIP pseudo label & 85.3 \\
AMST w/o CR & 84.9 \\
AMST w/o Multi-head & 86.1 \\
DreamDA & 86.5 \\
\bottomrule
\end{tabular}}
\caption{Comparison between AMST and standard pseudo label approaches as well as ablation on the influence of CR and multi-head training.}
\label{ablation_AMST}
\end{minipage}
\end{minipage}





\begin{wrapfigure}{r}{0.3\columnwidth}
\setlength{\belowcaptionskip}{-0.5cm}
\centering
\includegraphics[width=0.3\columnwidth]{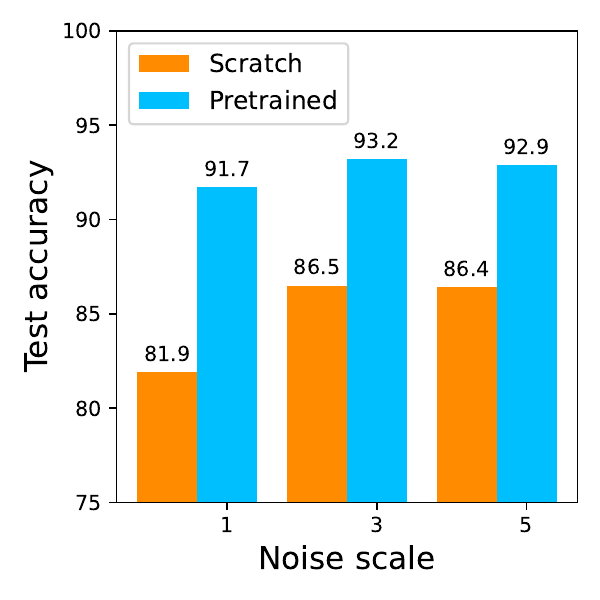}
    \caption{Ablation of different noise scales.}
    \label{ablation_noise_scale}
\end{wrapfigure}

\noindent\textbf{Latent Perturbation.} 
Here we study the effect of different perturbation strategies by comparing the following locations where Gaussian noise is injected: (1) \(x_0\). (2) \(\{x_t\}^{T-1}_1\). (3) U-Net decoder instead of the bottleneck for each intermediate step. Table~\ref{ablation_noise_inversion} shows our proposed latent perturbation at the U-Net bottleneck layer achieves the best performance. 
Notably, sub-optimal perturbation approaches lead to a significant decrease in accuracy, highlighting the challenge of discovering effective perturbation methods.
Additionally, we examine the effect of Gaussian noise with different scales, as shown in Fig.~\ref{ablation_noise_scale}. Our findings indicate that a noise scale of 3 is optimal among the tested values of \{1,3,5\}.




\noindent\textbf{AMST.} This section compares AMST with alternative pseudo labeling methods, and also studies the impact of two components within AMST, consistency regularization (CR) and multi-head training. Table.~\ref{ablation_AMST} indicates that AMST outperforms soft pseudo labeling~\cite{arazo2020pseudo} and CLIP pseudo labeling, which makes use of the pretrained CLIP~\cite{radford2021learning} model to predict labels. Furthermore, we in turn removed CR and multi-head training from AMST and found consistent decreases in accuracy, demonstrating the effectiveness of our unified framework.

\section{Conclusion}
We introduce DreamDA, a novel framework for generative data augmentation. Extensive experiments verify the effectiveness of our approach, emphasizing the potential of generative DA using large pre-trained models. In the future, we would like to investigate faster sampling techniques~\cite{meng2023distillation} to reduce the computational cost of our method. Furthermore, ethical and societal issues should be considered carefully when adopting generative DA in real-life applications.

\clearpage  

%
%
\bibliographystyle{splncs04}
\bibliography{main}
\end{document}